\documentclass[12pt,halfline,a4paper]{ouparticle}

\usepackage{hyperref}
\usepackage{booktabs}
\usepackage{caption}
\usepackage[normalem]{ulem}
\useunder{\uline}{\ul}{}
\usepackage{lscape}

\usepackage{setspace} 
\usepackage{etoolbox}
\AtBeginEnvironment{quote}{\par\singlespacing\small}
\AtBeginEnvironment{tabular}{\footnotesize}

\begin{document}

\title{Automated Paper Screening for Clinical Reviews\break {Using Large Language Models}}

\author{%
\name{Eddie Guo$^{1,*}$, Mehul Gupta$^1$, Jiawen Deng$^2$, Ye-Jean Park$^2$, Mike Paget$^1$, Christopher Naugler$^1$}
\address{\small$^1$Cumming School of Medicine, University of Calgary, Calgary, Canada}
\address{\small$^2$Temerty Faculty of Medicine, University of Toronto, Toronto, Canada}
\email{$^*$Corresponding author: eddie.guo@ucalgary.ca, 1-587-988-0292, 3330 Hospital Dr NW, Calgary, AB, Canada T2N 4N1}
}

\abstract{
\textbf{Objective} To assess the performance of the OpenAI GPT API in accurately and efficiently identifying relevant titles and abstracts from real-world clinical review datasets and compare its performance against ground truth labelling by two independent human reviewers.

\textbf{Methods} We introduce a novel workflow using the OpenAI GPT API for screening titles and abstracts in clinical reviews. A Python script was created to make calls to the GPT API with the screening criteria in natural language and a corpus of title and abstract datasets that have been filtered by a minimum of two human reviewers. We compared the performance of our model against human-reviewed papers across six review papers, screening over 24,000 titles and abstracts.

\textbf{Results} Our results show an accuracy of 0.91, a sensitivity of excluded papers of 0.91, and a sensitivity of included papers of 0.76. On a randomly selected subset of papers, the GPT API demonstrated the ability to provide reasoning for its decisions and corrected its initial decision upon being asked to explain its reasoning for a subset of incorrect classifications.

\textbf{Conclusion} The GPT API has the potential to streamline the clinical review process, save valuable time and effort for researchers, and contribute to the overall quality of clinical reviews. By prioritizing the workflow and acting as an aid rather than a replacement for researchers and reviewers, the GPT API can enhance efficiency and lead to more accurate and reliable conclusions in medical research.
}

\date{\today}


\maketitle

\section{BACKGROUND AND SIGNIFICANCE}
\label{sec:intro}

Knowledge synthesis, the process of integrating and summarizing relevant studies in the literature to gain an improved understanding of a topic, is a key component in identifying knowledge gaps and informing future research endeavors for a topic of interest \cite{Sargeant2020-pd,Garritty2019-eo}. Systematic and scoping reviews are some of the most commonly used and rigorous forms of knowledge synthesis across multiple disciplines \cite{Sargeant2020-pd,Garritty2019-eo}. Given that the results from systematic and scoping reviews can inform guidelines, protocols, and decision-making processes, particularly for stakeholders in the realms of healthcare, the quality of the evidence presented by such reviews can have significant impacts \cite{Luchini2021-oo}.

The quality of systematic and scoping reviews is highly dependent on the comprehensiveness of the database searches and the subsequent article screening processes. Overlooking relevant articles during these critical steps can lead to bias \cite{Gartlehner2020-eg}, while including discrepant studies can yield misleading conclusions and increase discordant heterogeneity \cite{Fletcher2007-fm}. Thus, guidelines surrounding the conduct of clinical reviews, such as the Cochrane Handbook \cite{Higgins2021-zn}, recommend that article screening should be completed in duplicate by at least two independent reviewers.

However, duplicate screening effectively doubles the financial and human resources needed to complete systematic reviews compared to single screening. This is especially problematic for small research groups, review projects with broad inclusion criteria (such as network meta-analyses), or time-constrained review projects (such as reviews relating to COVID-19 during the early stages of the pandemic) \cite{Chai2021-oc}. Additionally, there is often substantial inter-rater variability in screening decisions, leading to additional time spent on discussions to resolve disagreements \cite{Tuijn2012-pu}. Due to the time constraints and wasted resources that are often features of duplicate screening, research studies may also include a more tailored, sensitive search strategy that can lead to missing several articles during the retrieval process \cite{Rathbone2015-ll}. Furthermore, although the nuances of each study differ, many systematic reviews may contain thousands of retrieved articles, only to exclude the majority (i.e. up to $\sim$90\%) from the title and abstract screening \cite{Rathbone2015-ll,Polanin2019-zk}.

Recent developments in artificial intelligence and machine learning (ML) have made it possible to semi-automate or fully automate repetitive steps within the systematic review workflow \cite{Marshall2019-ka,Marshall2023-ob,Blaizot2022-al}. Prominent examples of such applications include RobotReviewer and TrialStreamer \cite{Marshall2017-pi,Marshall2020-yo}, which are ML models developed to extract information from scientific articles or abstracts to judge study quality and infer treatment effects. More specifically, RobotReviewer (2016) was shown to have similar capabilities to assess the risk of bias assessment as a human reviewer, only differing by around 7\% in accuracy \cite{Marshall2016-rc}. Similarly, TrialStreamer was a system developed to extract key elements of information from full texts, such as inferring which interventions in a clinical paper worked best, along with comparisons in study outcomes between all relevant extracted full-texts of a topic indexed on MEDLINE \cite{Nye2020-gc}.

While there have been previous attempts at automating the title and abstract screening process, they often involved labor- or computationally-intensive labeling, pre-training, or vectorizations \cite{Moreno-Garcia2023-ej}. For instance, Rayyan and Abstracker are two free Web tools that provide a semi-automated approach to article filtering by using natural language processing algorithms to learn when and where a reviewer includes or excludes an article, and subsequently mimicking a similar approach \cite{Wallace2010-ug,Ouzzani2016-jl}. Rayyan also demonstrated high specificity wherein 98\% of all relevant articles were included after the tool had screened 75\% of all articles to be analyzed in a study \cite{Olofsson2017-uo}. While automation using these tools was found to save time, there was still quite minimal to substantive risk that there would be missing studies if the tool was fully independent/automated \cite{Wallace2010-ug,Ouzzani2016-jl}. Furthermore, current programs may use previously standard methods, including n-grams, in comparison to more updated techniques, such as the generative pre-trained transformer (GPT) model that is trained with data from a general domain and does not require additional training to learn embeddings that can represent the semantics and contexts of words in relation to other words \cite{Shree2020-rf,OMara-Eves2015-qt}.

In this paper, we introduce a novel workflow to screen titles and abstracts for clinical reviews by providing plain language prompts to the publicly available OpenAI GPT API. We aimed to assess GPT models’ ability to accurately and efficiently identify relevant titles and abstracts from real-world clinical review datasets, as well as their ability to explain their decisions and reflect on incorrect classifications. We compare the performance of our model against ground truth labeling by two independent human reviewers across six review papers in the screening of over 24,000 titles and abstracts.

\begin{figure}[!t]
    \centering
    \includegraphics[width=\columnwidth]{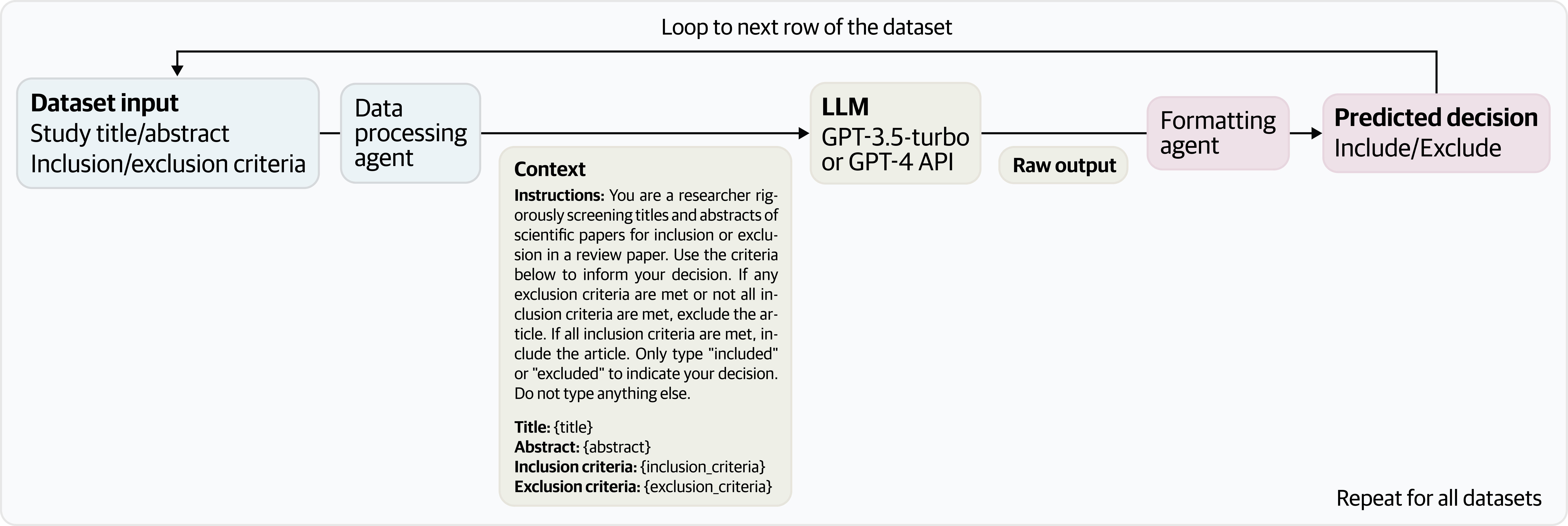}
    \caption{Overview of the Python script to automate screening with the GPT API.}
    \label{fig:pipeline}
\end{figure}

\section{METHODS}
\label{sec:methods}

In our study, we obtained a corpus of title and abstract datasets that have already been filtered by a minimum of two human reviewers to train our model. Subsequently, we created a Python script that provides the screening criteria for each paper to the OpenAI GPT API. We then passed each paper to the API using a consistent instruction prompt to determine whether a paper should be included or excluded based on the contents of its title and abstract. The overall accuracy, sensitivity of both included and excluded papers, and inter-rater reliability via Cohen’s kappa were computed against the human-reviewed papers. All data and code are available in Mendeley Datasets at the following doi: 10.17632/np79tmhkh5.1.

\subsection{Data Collection}
\label{subsec:data-collection}

To validate our proposed inclusion/exclusion methodology, we obtained a total of six title/abstract screening datasets from different systematic and scoping review projects. These projects cover a diverse range of medical science topics and vary in size, methodology, and complexity of screening criteria (Table \ref{tab:dataset-info} and Supplementary Table 1). We obtained the inclusion/exclusion decision from expert reviewers for each title/abstract entry, as well as the criteria provided to the expert reviewers during the screening process. A summary of the review characteristics is presented in Table \ref{tab:dataset-requirements}.

\begin{table}[!th]
    \centering
    \caption{Included studies and their characteristics. The first five datasets are systematic reviews with meta-analyses. The last study is a scoping review.}
    \label{tab:dataset-info}
    \begin{tabular}{p{5cm}p{2cm}p{1.25cm}p{3cm}p{2.5cm}}
        \midrule
        Study Title [doi] & Dataset Name & n (n included) & Study Type & Study Topic \\
        \midrule
        Efficacy and safety of ivermectin for the treatment of COVID-19: a systematic review and meta-analysis [10.1093/qjmed/hcab247] & IVM & 279 (35) & Systematic review and meta-analysis of randomized and non-randomized trials & COVID-19 Treatment, Antimalarials \\
        \midrule
        Efficacy and safety of selective serotonin reuptake inhibitors in COVID-19 management: a systematic review and meta-analysis [10.1016/j.cmi.2023.01.010] & SSRI & 3,989 (29) & Systematic review and meta-analysis of randomized and non-randomized trials & COVID-19 Treatment, Antidepressants \\
        \midrule
        Efficacy of lopinavir–ritonavir combination therapy for the treatment of hospitalized COVID-19 patients: a meta-analysis [10.2217/fvl-2021-0066] & LPVR & 1,456 (91) & Systematic review and meta-analysis of randomized and non-randomized trials & COVID-19 Treatment, Antiretrovirals \\
        \midrule
        The use of acupuncture in patients with Raynaud’s syndrome: A systematic review and meta-analysis of randomized controlled trials [10.1177/09645284221076504] & RAYNAUDS & 942 (6) & Systematic review and meta-analysis of randomized controlled trials & Raynaud’s Syndrome, Acupuncture \\
        \midrule
        Comparative efficacy of adjuvant non-opioid analgesia in adult cardiac surgical patients: A network meta-analysis [10.1053/j.jvca.2023.03.018] & NOA & 14,771 (354) & Systematic review and network meta-analysis of randomized controlled trials & Post-Operative Pain, Analgesics \\
        \midrule
        Assessing the research landscape and utility of LLMs in the clinical setting: protocol for a scoping review$^*$ & LLM & 2,870 (23) & Scoping review & Machine Learning in Clinical Medicine \\
        \midrule
        \textbf{Total} & - &  24,307 (538) & - & - \\
        \midrule
        \multicolumn{5}{l}{\footnotesize$^*$Registered with OSF, not yet published: \href{https://osf.io/498k6}{https://osf.io/498k6}.}
    \end{tabular}
\end{table}

\subsection{App Creation}
\label{subsec:app-creation}
Given a dataset, \verb|df_info|, containing information about inclusion and exclusion criteria of the datasets containing titles and abstracts to be reviewed, the app calls the OpenAI GPT API to classify each paper to be screened as either included or excluded. The app was coded in Python. The prompt given to the GPT API was as follows:

\begin{quote}
    \textbf{Instructions:} You are a researcher rigorously screening titles and abstracts of scientific papers for inclusion or exclusion in a review paper. Use the criteria below to inform your decision. If any exclusion criteria are met or not all inclusion criteria are met, exclude the article. If all inclusion criteria are met, include the article. Only type "included" or "excluded" to indicate your decision. Do not type anything else.

    \textbf{Abstract:} \{abstract\}

    \textbf{Inclusion criteria:} \{inclusion\_criteria\}

    \textbf{Exclusion criteria:} \{exclusion\_criteria\}

    \textbf{Decision:}
\end{quote}

Where ``\textbf{Decision:}'' is whether GPT API includes or excludes the article. Thus, the algorithm is as follows:

\begin{verbatim}
data_df <- load(df_info)
for each dataset in data_df:
    for each row in dataset:
        prompt <- instructions + title + abstract + inclusion criteria \
            + exclusion criteria
        decision <- GPT(prompt)
        row[‘decision’] <- decision
    save(dataset)
\end{verbatim}

\subsection{Assessment and Data Analysis}
\label{subsec:assessment}

After the app was run on all datasets included in our analysis, the following metrics were computed: absolute agreement, sensitivity for decision tags, and a classification report returned by scikit-learn.

A subset of the results was selected for GPT to explain its reasoning. The following prompt was appended to the beginning of the original prompt given to the API: “Explain your reasoning for the decision given with the information below.” The human and GPT decisions were appended to the end of the prompt. A subset of incorrect results was selected for GPT to reflect on its incorrect answers. The following prompt was appended to the beginning of the original prompt given to the API: “Explain your reasoning for why the decision given was incorrect with the information below.” The human and GPT decisions were appended to the end of the prompt. The results are tabulated in Table \ref{tab:reasoning}.

\begin{table}[!t]
    \centering
    \caption{Data formatting for the Python script automating screening with the GPT API. All non-English characters were removed prior to analysis.}
    \label{tab:dataset-requirements}
    \begin{tabular}{ll}
        \midrule
        Data                 & Columns  \\
        \midrule
        \verb|df_info|    & \begin{tabular}[c]{@{}l@{}}\verb|Dataset Name| (str): name of dataset\\ \verb|Inclusion Criteria| (str): screening inclusion criteria\\ \verb|Excusion Criteria| (str): screening exclusion criteria\end{tabular} \\
        \midrule
        \verb|dataset|$^*$ & \begin{tabular}[c]{@{}l@{}}\verb|title| (str): paper title\\ \verb|abstract| (str): paper abstract \end{tabular} \\
        \midrule
        \multicolumn{2}{l}{\footnotesize$^*$The name of \texttt{dataset} must match \texttt{Dataset Name} in \texttt{df\_info}.}
    \end{tabular}
\end{table}

\section{Results}
\label{sec:results}

\begin{figure}[!th]
    \centering
    \includegraphics[width=\columnwidth]{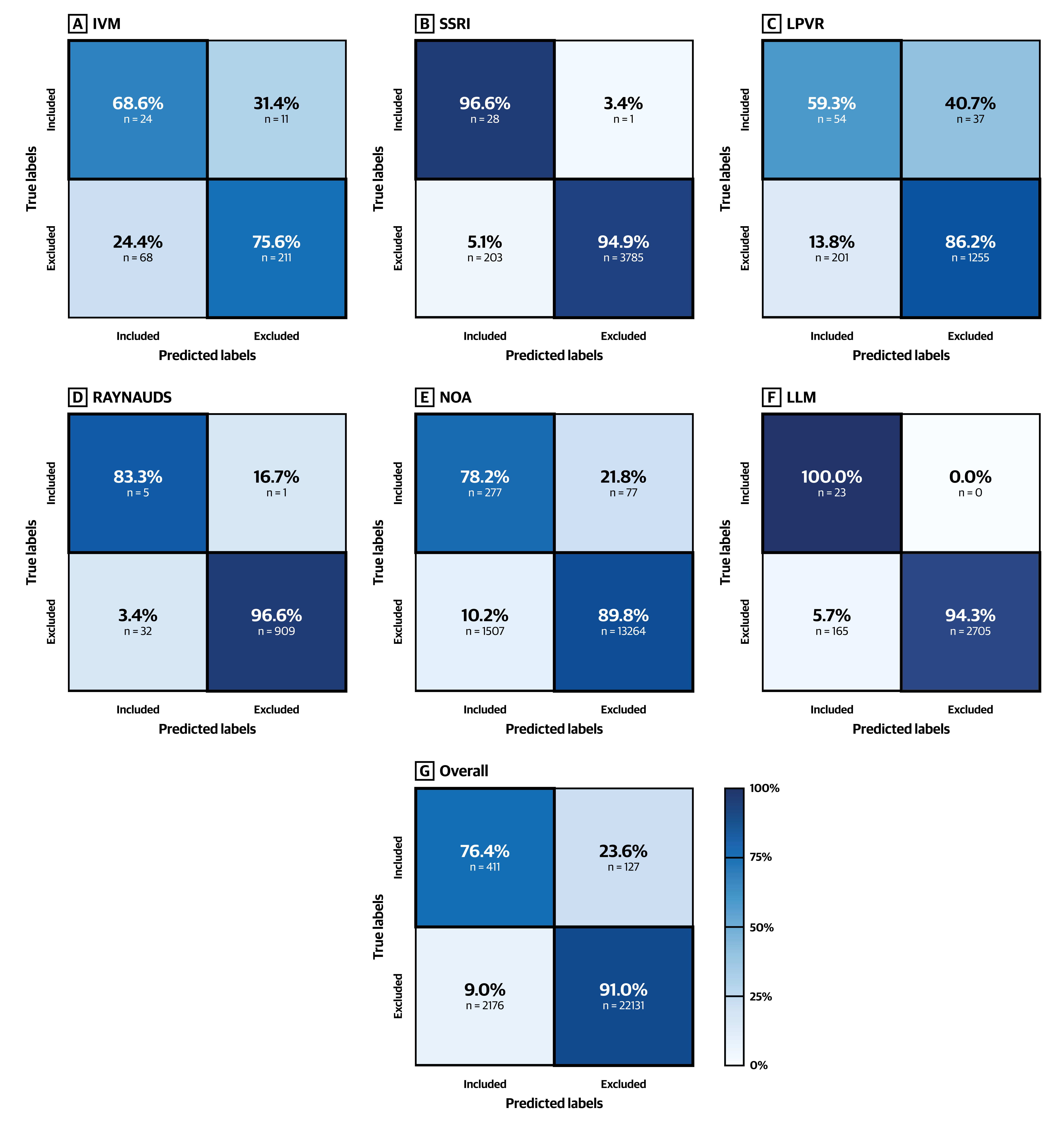}
    \caption{Confusion matrices for the included and excluded papers for all datasets (A-F) and the overall performance (G).}
    \label{fig:confusion-matrix}
\end{figure}

The overall accuracy of GPT was 0.91, the sensitivity of included papers was 0.76, and the sensitivity of excluded papers was 0.91 (Table [results], Fig. [confusion matrix]). On the NOA dataset (n=14,771 with 354 included abstracts), the model ran for 643m 50.8s with an approximate cost of 25 USD. The dataset characteristics are detailed in Table \ref{tab:dataset-info}, the model performance is in Table \ref{tab:results} and visualized in Fig. \ref{fig:confusion-matrix}, and the reasoning from GPT is tabulated in Table \ref{tab:reasoning}.

\begin{table}[!tbh]
    \centering
    \caption{Performance of GPT in screening titles and abstracts against a human reviewer ground truth. Kappa (Human) is the agreement between two independent human reviewers. Kappa (Screen) is the agreement between GPT and the final papers included/excluded in each dataset.}
    \label{tab:results}
    \begin{tabular}{llp{1.75cm}p{1.75cm}p{1.5cm}p{1.5cm}}
        \midrule
        Dataset                    & Accuracy & Sensitivity   (Included) & Sensitivity   (Excluded) & Kappa   (Human) & Kappa   (Screen) \\
        \midrule
        IVM                        & 0.748    & 0.686                    & 0.756                    & 0.72            & 0.26             \\
        SSRI                       & 0.846    & 0.966                    & 0.949                    & 0.58            & 0.21             \\
        LPVR                       & 0.949    & 0.593                    & 0.862                    & 0.51            & 0.25             \\
        RAYNAUDS                   & 0.965    & 0.833                    & 0.966                    & 0.91            & 0.22             \\
        NOA                        & 0.895    & 0.782                    & 0.898                    & 0.35            & 0.23             \\
        LLM                        & 0.943    & 1.000                    & 0.942                    & 0.69            & 0.21             \\
        \midrule
        Total   (Weighted Average) & 0.907    & 0.764                    & 0.910                     & -               & -               \\
        \midrule
    \end{tabular}
\end{table}

\begin{landscape}
\begin{table}[]
\centering
\caption{Decision and reasoning of the GPT API for a subset of four example titles and abstracts. Full abstracts are in IVM (Table \ref{tab:dataset-info}).}
\label{tab:reasoning}
\begin{tabular}{@{}ll@{}}
\toprule
Prompt & Decision and Reasoning \\ \midrule
\begin{tabular}[c]{p{11cm}}\textbf{Title:} Use of ivermectin in the treatment of Covid-19: A pilot trial\\ \textbf{Abstract:} Objectives: In this randomized open-label trial pilot study we assessed the antiviral effects and safety of various doses of ivermectin in patients with mild clinical symptoms of COVID-19. Method(s): Patients were randomly assigned to receive standard of care (SOC) treatment at hospital admission...\end{tabular} & \begin{tabular}[c]{p{12cm}}\textbf{Human:} Included\\ \textbf{GPT:} Included\\ \textbf{Reasoning:} The study meets all of the inclusion criteria. It is a randomized controlled trial that compares ivermectin against standard of care treatment in adult COVID-19 patients. The study reports outcomes related to time to viral clearance and adverse events. Additionally, the study is published in a peer-reviewed journal. Therefore, the article should be included in the review paper.\end{tabular} \\
\midrule
\begin{tabular}[c]{p{11cm}}\textbf{Title:} What does the COVID-19 pandemic mean for the next decade of onchocerciasis control and elimination?\\ \textbf{Abstract:} Background: Mass drug administration (MDA) of ivermectin for onchocerciasis has been disrupted by the coronavirus disease 2019 (COVID-19) pandemic. Mathematical modelling can help predict how missed/delayed MDA will affect short-term epidemiological trends and elimination prospects by 2030...\end{tabular} & \begin{tabular}[c]{p{12cm}}\textbf{Human:} Excluded\\ \textbf{GPT:} Excluded\\ \textbf{Reasoning:} The article does not meet any of the inclusion criteria or exclusion criteria listed. It is not a randomized controlled trial or comparative cohort study, it does not compare ivermectin against standard of care or control group/placebo, it does not include adult COVID-19 inpatients and/or outpatients, and it does not report any of the specified outcomes. Additionally, there is no indication that the article is non-peer-reviewed. Therefore, the article is excluded.\end{tabular} \\
\midrule
\begin{tabular}[c]{p{11cm}}\textbf{Title:} Use of Ivermectin Is Associated With Lower Mortality in Hospitalized Patients With Coronavirus Disease 2019: The Ivermectin in COVID Nineteen Study\\\textbf{Abstract:} Background: Ivermectin was shown to inhibit severe acute respiratory syndrome coronavirus 2 replication in vitro, which has led to off-label use, but clinical efficacy has not been described previously. Research Question: Does ivermectin benefit hospitalized coronavirus disease 2019 (COVID-19) patients?...\end{tabular} & \begin{tabular}[c]{p{12cm}}\textbf{Human:} Included\\ \textbf{GPT:} Excluded\\ \textbf{Reasoning:} The article meets all of the inclusion criteria and none of the exclusion criteria, therefore it should be included in the review paper.\\ \textbf{Reflection:} The given decision of excluding the article is incorrect. The article meets all the inclusion criteria as it is a non-randomized comparative cohort study that compares ivermectin against standard of care or control group/placebo, includes adult COVID-19 inpatients, and reports mortality incidence. The article does not use ivermectin for prophylaxis of COVID-19 and is a peer-reviewed article. Therefore, the correct decision should be "Included".\end{tabular} \\
\midrule
\begin{tabular}[c]{p{11cm}}\textbf{Title:} What to expect from different drugs used in the treatment of COVID-19: A study on applications and in vivo and in vitro results\\ \textbf{Abstract:} The end of 2019 was marked by the emergence of a new type of coronavirus (SARS-CoV-2), which has killed more than 240,000 people...\end{tabular} & \begin{tabular}[c]{p{12cm}}\textbf{Human:} Excluded\\ \textbf{GPT:} Included\\ \textbf{Reasoning:} The study is not a randomized controlled trial or a non-randomized comparative cohort study, which is an inclusion criteria. Therefore, the article does not meet all inclusion criteria and must be excluded.\\ \textbf{Reflection:} As per the given inclusion and exclusion criteria, the decision given is incorrect. The abstract does not mention the use of ivermectin in any of the studies. Therefore, the article cannot be included based on the inclusion criteria.\end{tabular} \\ \bottomrule
\end{tabular}
\end{table}
\end{landscape}

\section{DISCUSSION}
\label{sec:discussion}

In this study, we assessed the performance of the OpenAI GPT API in the context of clinical review paper inclusion/exclusion criteria selection. The data showed that the overall accuracy of the app was 0.91, indicating a high level of agreement between the app's decisions and the reference standard. The sensitivity of included papers was 0.76, suggesting that the app had moderate performance in correctly identifying relevant papers (Table \ref{tab:results}, Fig. \ref{fig:confusion-matrix}). The sensitivity of excluded papers was 0.91, demonstrating that the app was effective in excluding irrelevant papers. These results highlight the potential of the GPT API to support the clinical review process.

\subsection{Implications of GPT API's Performance in the Review Process}

The GPT API’s performance has several implications for the efficiency and consistency of clinical review paper inclusion/exclusion criteria selection. By prioritizing the workflow and acting as an aid rather than a replacement for researchers and reviewers, the GPT API has the potential to streamline the review process. This enhanced efficiency could save valuable time and effort for researchers and clinicians, allowing them to focus on more complex tasks and in-depth analysis. Further, the API does not require pretraining on the user's end and can provide reasoning for its decision to either include or exclude papers, an aspect traditional natural language processing algorithms lack in automated or semi-automated paper screening (Table \ref{tab:reasoning}). Interestingly, upon being asked to explain its reasoning for a subset of incorrect classifications, the GPT API corrected its initial decision. Ultimately, this increased efficiency paired with reasoning capabilities could contribute to the overall quality of clinical reviews, leading to more accurate and reliable conclusions in medical research.

The use of the GPT API in the review process could also promote consistency in the selection of relevant papers. By automating certain aspects of the process and acting as an aid to researchers and clinicians, the API can streamline the review process and help reduce the potential for human error and bias, leading to more objective and reliable results \cite{Zhang2022-rq}. This increased consistency could, in turn, improve the overall quality of the evidence synthesized in clinical reviews, providing a more robust foundation for medical decision-making and the development of clinical guidelines.

The GPT API's potential as a decision tool becomes particularly valuable when resources are limited. In such situations, the API can be employed as a first-pass decision aid, streamlining the review process and allowing human screeners to focus on a smaller, more relevant subset of papers. By automating the initial screening process, the API can help reduce the workload for researchers and clinicians, enabling them to allocate their time and effort more efficiently.

Using the GPT API as a first-pass decision aid can also help mitigate the risk of human error and bias in the initial screening phase, promoting a more objective and consistent selection of papers. While the API's sensitivity for including relevant papers may not be perfect, its high specificity for excluding irrelevant papers can still provide valuable support in narrowing down the pool of potentially relevant studies \cite{Rathbone2015-ll}. This can be particularly beneficial in situations where a large number of papers need to be screened, and human resources are scarce \cite{Van_de_Schoot2021-ag}.

\subsection{Limitations and Challenges in Implementing GPT API in the Review Process}

While the GPT API shows promise in streamlining the review process, it is important to acknowledge its limitations and challenges. One notable limitation is the disparity between the high specificity of 0.91 for excluding papers and the lower sensitivity of 0.76 for including papers. This discrepancy suggests that while the API is effective in excluding irrelevant papers, it may not be as proficient in identifying relevant papers for inclusion. This could lead to the omission of important studies in the review process, potentially affecting the comprehensiveness and quality of the final review.

Therefore, the GPT API should not be considered a replacement for human expertise. Instead, it should be viewed as a complementary tool that can enhance the efficiency and consistency of the review process. Human screeners should still be involved in the final decision-making process, particularly in cases where the API's sensitivity for including relevant papers may be insufficient \cite{Chai2021-oc}. By combining the strengths of the GPT API with human expertise, researchers can optimize the review process and ensure the accuracy and comprehensiveness of the final review.

\subsection{Future Research and Development}

Several avenues for future research and development include refining the GPT API's performance in the clinical review paper context, incorporating meta-data such as study type and year, and exploring few-shot learning approaches. Additionally, training a generator-discriminator model via fine-tuning could improve the API's performance \cite{noauthor_undated-un}. Expanding the application of the GPT API to other areas of medical research or literature review could also be explored. This would involve large language models (LLMs) for tasks such as identifying and extracting study design information, patient characteristics, and adverse events. As the maximum token length increases with future iterations of the GPT model, screening entire papers may become feasible \cite{noauthor_undated-gq}. Furthermore, exploring the use of LLMs to generate clinical review papers could be a promising research direction.

\section{CONCLUSION}
\label{sec:conclusion}

The GPT API shows potential as a valuable tool for improving the efficiency and consistency of clinical review paper inclusion/exclusion criteria selection. While there are limitations and challenges to its implementation, its performance in this study suggests that it could have a broader impact on clinical review paper writing and medical research. Future research and development should focus on refining the API's performance, expanding its applications, and exploring its potential in other aspects of clinical research.

\section*{ACKNOWLEDGEMENTS}
\label{sec:acknowledgements}

We would like to acknowledge the following expert reviewers for providing the screening decisions in the review datasets used in this study and for agreeing to make the datasets publicly available: Abhinav Pillai, Mike Paget, Christopher Naugler, Kiyan Heybati, Fangwen Zhou, Myron Moskalyk, Saif Ali, Chi Yi Wong, Wenteng Hou, Umaima Abbas, Qi Kang Zuo, Emma Huang, Daniel Rayner, Cristian Garcia, Harikrishnaa Ba Ramaraju, Oswin Chang, Zachary Silver, Thanansayan Dhivagaran, Elena Zheng, Shayan Heybati.

\section*{CONFLICT OF INTEREST}
\label{sec:coi}

The authors declare that there is no conflict of interest.

\section*{FUNDING}
\label{sec:funding}

This study was not funded.

\section*{CREDIT AUTHOR STATEMENT}
\label{sec:credit}

\textbf{Eddie Guo:} Conceptualization, methodology, software, formal analysis, investigation, writing - original draft, writing - review \& editing, visualization, supervision, project administration; \textbf{Mehul Gupta:} Conceptualization, methodology, investigation, writing - original draft, writing - review \& editing, supervision, project administration; \textbf{Jiawen Deng:} Methodology, software, formal analysis, investigation, data curation, writing - original draft, visualization; \textbf{Ye-Jean Park:} Methodology, formal analysis, investigation, data curation, writing - original draft, visualization; \textbf{Mike Paget:} Writing - reviewing \& editing; \textbf{Christopher Naugler:} Writing - review \& editing

\bibliographystyle{jamia}
\bibliography{references}

\end{document}